\documentclass[acmtog, nonacm]{acmart} 

\acmSubmissionID{192}

\usepackage{xcolor}
\definecolor{darkred}{rgb}{0.75, 0.1, 0.1}

\usepackage{amsmath} % For mathematical symbols
\usepackage{algorithm} 
\usepackage{algpseudocode} 
\usepackage{svg}
\usepackage{tabularx} 
\usepackage{stfloats} 

\makeatletter
    \renewcommand{\@mkauthorsaddresses}{}
\makeatother

\setcopyright{cc}
\copyrightyear{2025}
\acmYear{}
\acmDOI{}
\acmConference[SIGGRAPH]{}{August 10--14, 2025}{Vancouver, BC}
\acmISBN{}

\newcommand{\TODO}[1]{} % don't print TODOs when compiling the paper for submission

\acmJournal{TOG}

\begin{document}

\title{SqueezeMe: Mobile-Ready Distillation of Gaussian Full-Body Avatars}

\author{Forrest Iandola}
\affiliation{\institution{Meta Reality Labs} \country{USA}}
\author{Stanislav Pidhorskyi}
\affiliation{\institution{Meta Codec Avatars Lab} \country{USA}}
\author{Igor Santesteban}
\affiliation{\institution{Meta Codec Avatars Lab} \country{USA}}
\author{Divam Gupta}
\affiliation{\institution{Meta Codec Avatars Lab} \country{USA}}
\author{Anuj Pahuja}
\affiliation{\institution{Meta Codec Avatars Lab} \country{USA}}
\author{Nemanja Bartolovic}
\affiliation{\institution{Meta Codec Avatars Lab} \country{Switzerland}}
\author{Frank Yu}
\affiliation{\institution{Meta Codec Avatars Lab} \country{USA}}
\author{Emanuel Garbin}
\affiliation{\institution{Meta Codec Avatars Lab} \country{Israel}}
\author{Tomas Simon}
\affiliation{\institution{Meta Codec Avatars Lab} \country{USA}}
\author{Shunsuke Saito}
\affiliation{\institution{Meta Codec Avatars Lab} \country{USA}}

\renewcommand{\shortauthors}{Iandola et al.}

\begin{abstract}
Gaussian-based human avatars have achieved an unprecedented level of visual fidelity. However, existing approaches based on high-capacity neural networks typically require a desktop GPU to achieve real-time performance for a single avatar, and it remains non-trivial to animate and render such avatars on mobile devices including a standalone VR headset due to substantially limited memory and computational bandwidth.
In this paper, we present SqueezeMe, a simple and highly effective framework to convert high-fidelity 3D Gaussian full-body avatars into a lightweight representation that supports both animation and rendering with mobile-grade compute. 
Our key observation is that the decoding of pose-dependent Gaussian attributes from a neural network creates non-negligible memory and computational overhead. 
Inspired by blendshapes and linear pose correctives widely used in Computer Graphics, we address this by distilling the pose correctives learned with neural networks into linear layers.
Moreover, we further reduce the parameters by sharing the correctives among nearby Gaussians.
Combining them with a custom splatting pipeline based on Vulkan, we achieve, for the first time, simultaneous animation and rendering of 3 Gaussian avatars in real-time (72 FPS) on a Meta Quest 3 VR headset. 
Demo videos are available at \url{https://forresti.github.io/squeezeme}.
\end{abstract}




\begin{teaserfigure}
    % we used linearized + 4k correctives
    \centering
    \includegraphics[width=0.9\textwidth]{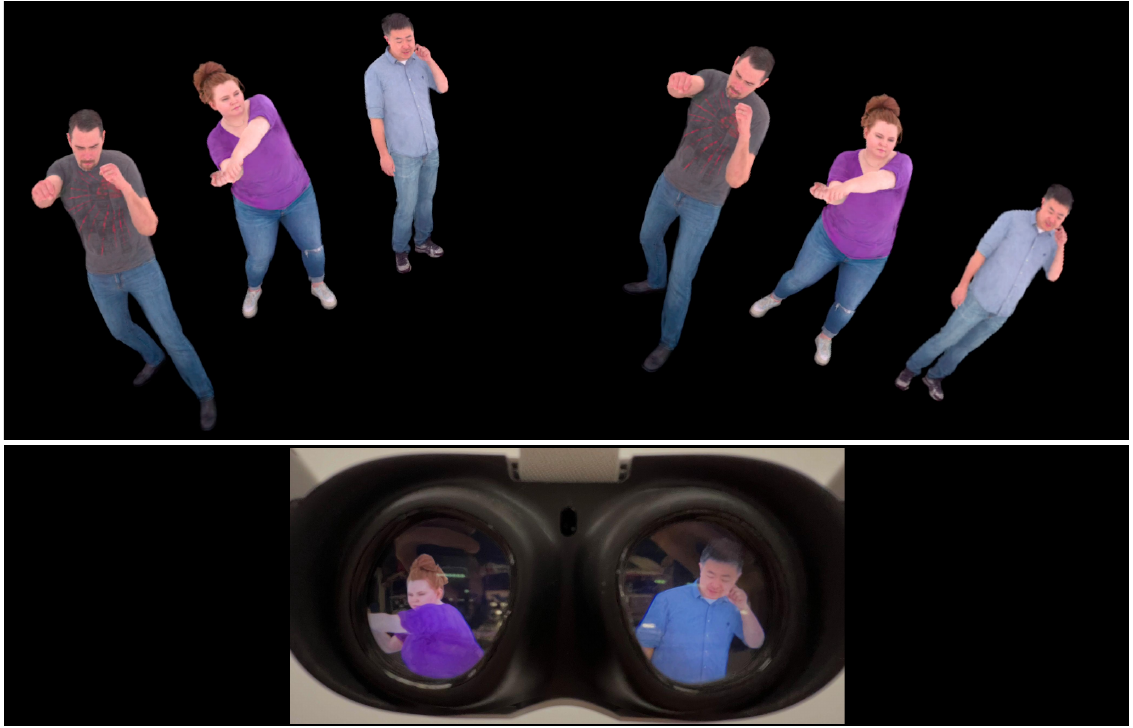}
    \caption{\textbf{With SqueezeMe, we simultaneously run 3 Gaussian avatars locally on a Meta Quest 3 VR headset.} Upper: Stereo view in VR. Lower: View into the VR headset lenses.}
    \Description{3 avatars}
    \label{fig:teaser}
\end{teaserfigure}

\maketitle

\section{Introduction}

Augmented Reality (AR) and Virtual Reality (VR) have shown great promise in delivering immersive experiences that blur the lines between the physical and virtual worlds.
At the heart of these experiences lies photorealistic full-body avatars, whose animation and appearance are indistinguishable from reality. 
Such avatars enable a wide range of applications, including telepresence, virtual try-on, and immersive gaming.
To make such technology accessible to a broader population, standalone AR/VR devices with lightweight form factors are required, necessitating a highly restrictive computational budget. 
Our goal is to enable the animation and rendering of photorealistic avatars on mobile-grade compute, thereby paving the way for widespread adoption of authentic digital humans in AR/VR.

Recent advances in accelerating the rendering of photorealistic 3D scenes based on NeRF~\cite{chen2023mobilenerf,reiser2023merf,reiser2024binary} and 3D Gaussian Splatting (3DGS)~\cite{3DGS,Niedermayr_2024_CVPR} have shown promising results. 
Moreover, real-time rendering of volumetric videos on mobile devices has become possible~\cite{wang2024v}. 
However, these approaches are not directly applicable to avatars, because avatars require support for not only static scenes or the playback of pre-recorded sequences, but also novel animations computed on-the-fly for real-time driving of avatars~\cite{bagautdinov2021driving}.
Existing approaches that can achieve photorealism rely on high-capacity neural networks to decode non-rigid correctives given a driving signal such as body pose parameters~\cite{AnimatableGaussians,ARAH,HAHA}. 
Running such a decoder for every frame creates non-negligible memory and computational overhead, hindering the deployment of photorealistic avatars on device.

To address this, we present SqueezeMe, a novel approach to distilling photorealistic avatars based on 3D Gaussians into a lightweight representation that is ready for animation and rendering on mobile devices. 
Inspired by pose-dependent linear correctives, widely used for computer graphics~\cite{SMPL}, we compute a linear mapping from pose parameters to geometry and appearance parameters of 3D Gaussians, including rotation, displacements, scale, and spherical harmonics coefficients. 
More specifically, we first train 3D Gaussian avatars with pose-dependent corrective parameters defined on a UV map using a high-capacity convolutional neural network (CNN), which achieves comparable performance with a state-of-the-art Gaussian avatar method~\cite{AnimatableGaussians} with $5$ times fewer Gaussians.
Then, we extract key frames with associated Gaussian parameters to ensure the uniform coverage of various poses. 
We then solve a linear regression from the associated pose parameters to the target correctives. 

Although this linear distillation drastically simplifies the computations performed in the decoder, the size of the linear matrix remains relatively large, creating a non-trivial memory overhead for mobile compute. 
We observe that while static Gaussian parameters need to preserve high-frequency signals to be photorealistic, pose-dependent correctives tend to be low-frequency.
Based on this key insight, we further reduce the memory footprint by sharing the correctives among Gaussians that are adjacent on the UV map layout. 
This allows us to reduce the required memory roughly $16$ times, while minimally compromising visual fidelity and high-frequency person-specific details.

During inference, we render the Gaussians with the parameters computed from the linear model and use a custom renderer based on Vulkan.
Our experiments demonstrate that our approach allows us to animate and render up to 3 full-body avatars at real-time framerates (72FPS) on a Meta Quest 3 headset. We also carefully evaluate our approach and show that quality degradation by the proposed distillation is minimal.

In summary, our contributions include:
\begin{itemize}
    \item We present a framework to distill personalized Gaussian full-body avatars into a lightweight representation, enabling simultaneous animation and rendering of 3 avatars in real-time on a standalone VR headset (Quest 3) for the first time. 
    \item We learn a linear mapping from pose parameters to Gaussian corrective values using key frames of a Gaussian avatar based on high-capacity CNN.
    \item We further reduce the memory footprint by sharing pose-dependent correctives in the UV space with minimal quality degradation.
\end{itemize}

\section{Related Work}

\subsection{Photorealistic Full-Body Avatars}
Creating photrealistic avatars has been a long-standing problem in computer graphics and vision. Earlier works attempt modeling clothed humans by deforming a single template mesh such as SMPL~\cite{Loper2015SMPLAS} with texture maps~\cite{alldieck2018video,alldieck2018detailed,bagautdinov2021driving}. While the mesh-based approaches work well for tight clothing and skin regions, hair is typically difficult to handle due to topological constraints and difficulty in handling opacity.
Avatars based on Signed Distance Fields (SDF)~\cite{saito2021scanimate,ARAH,Chen2023CVPR,Chen2024ECCV,jiang2022selfrecon, TriHuman_2024} can handle topology changes, but have difficulty handling intricate geometric details such as hair. 
Volumetric representations such as Neural Volumes~\cite{lombardi2019neural} or NeRF~\cite{mildenhall2021nerf} address the limitation of meshes by explicitly integrating opacity in volumetric rendering equations. 
These approaches are later expended to human modeling~\cite{weng2020vid2actor,Peng2021CVPR,Peng2021ICCV,Liu2021NeuralA,Su2021NeurIPS,Jiang2022NeuManNH,li2022tava,Weng2022CVPR}.
While these volumetric representations are typically slow to render due to expensive evaluation, follow-up works accelerate the rendering for real-time inference by utilizing primitive approximation of volumetric fields with voxels~\cite{lombardi2021mixture,Remelli2022SIGGRAPH,Chen2023NEURIPS} or point clouds~\cite{zheng2023pointavatar, Su2023NPCNP, Prokudin2023dynamic}. 
However, these approaches still struggle with modeling detailed shape and appearance.
More recently, 3D Gaussian Splatting (3DGS)~\cite{3DGS} achieves better trade-off for rendering speed and fidelity by approximating radiance fields with anisotropic Gaussians. 
Full-body avatars based on 3DGS achieve state-of-the-art performance~\cite{Hu2024CVPR,Zielonka2023ARXIV,AnimatableGaussians,Qian2024CVPR,Pang2024CVPR,GaussianAvatar}.
While these approaches based on 3DGS achieves real-time animation and rendering on desktop-grade GPUs, decoding pose correctives per frame remains prohibitively slow or intractable to run on mobile compute due to high memory and computation requirements.
In contrast, our work proposes the distillation of these high-capacity CNN decoders into lightweight linear layers with significant memory reduction by corrective sharing, enabling the animation and rendering of pose-dependent animatable Gaussian avatars.

\subsection{Efficient Volumetric Rendering}
Efficient rendering of radiance fields are also extensively studied. 
Earlier works utilize multiplane image (MPI) ~\cite{wizadwongsa2021nex} or sparse voxel grids~\cite{hedman2021baking} to drastically reduce the number of evaluations per ray. 
More recent works further reduce the evaluation time to once per ray by extracting the surface mesh~\cite{chen2023mobilenerf,reiser2023merf,yariv2023bakedsdf,duckworth2024smerf,guedon2024sugar}.
3DGS rendering can also be accelerated by integer quantization, vector quantization, and hash grids to reduce the size footprint of 3DGS as demonstrated in~\cite{fan2023lightgaussian,Niedermayr_2024_CVPR,lee2024compact}.
These approaches are only applicable to static scenes, and more recent works support efficient rendering of 4D dynamic scenes using triplane~\cite{wu2024tetrirf}, voxels~\cite{wang2024videorf,im4d_lin2023high}, depth peeling~\cite{xu20244k4d}, and dynamic 3D Gaussians~\cite{sun20243dgstream,yang2023gs4d,xu2024longvolcap}. However, they are either non-trivial to stream and render on mobile devices~\cite{xu20244k4d,sun20243dgstream,wu2024tetrirf} or prone to blurry results~\cite{wang2024videorf}.
V3~\cite{wang2024v} achieves mobile streaming of dynamic Gaussian splatting by converting dynamic Gaussians to 2D image sequences and applying video codec.
Unlike these methods, which support only playback of pre-recorded sequences, our approach compresses the basis of correctives and effectively supports novel animations of photorealistic avatars with on-the-fly driving signals with mobile-grade compute.

\subsection{Efficient Animatable Avatars}
Improving the efficiency of rendering and animation of photorealistic avatars is an active research topic in the community. 
Pixel Codec Avatars~\cite{ma2021pixel} presents a method to reduce the memory and computation overhead of head avatars by offloading view-dependent texture decoding to per-pixel computation via neural deferred rendering~\cite{thies2019deferred} with low-resolution latent codes and high-resolution static maps. 
MoRF utilizes neural deferred shading for full-body avatar animation and rendering~\cite{bashirov2024morf}. However, due to the limited compute on mobile, the framerate and resolution are limited to 30 FPS with $640\times640$ on Qualcomm Snapdragon. Unfortunately, this is insufficient for VR, where we need at least 72 FPS with $2K$ resolution for two eyes.
SplattingAvatar~\cite{shao2024splattingavatar} is a hybrid representation of Gaussian Splats embedded on a tracked mesh, achieving 30 FPS for a single avatar on iPhone 13. However, this approach does not account for pose-dependent correctives, limiting its animation fidelity.
The closest to our work are Gaussian Blendshapes~\cite{ma20243d,Yan_2025_WACV} and Gaussian Eigen Models~\cite{gaussian_eigen_models}, where the linear basis of Gaussian parameters in head avatars is extracted from pretrained CNN decoders.
However, we find that simply extracting linear basis is still not sufficient to meet memory and computational requirements on mobile. Moreover, unlike head only avatars, full-body avatars need to explicitly account for pose-dependent deformations during the distillation process.
In contrast, we present, for the first time, a complete system to enable real-time animation and rendering of full-body Gaussian avatars that meet requirements for VR rendering.

\section{Preliminary}

\subsection{Gaussian Splatting}
3D Gaussian Splatting~\cite{3DGS} is a powerful representation for computer graphics.
Each Gaussian is parameterized by several terms including rotation, translation $\mu$, scale $\sigma$, and a set of spherical harmonics terms for view-dependent color.
The Gaussian's rotation and scale construct a covariance matrix $\Sigma$.
Each Gaussian also has a density term $\delta$, which defines the opacity at the center of the Gaussian. 

Let us view the Gaussians from a specific camera $c$, defined with focal length $(f_x, f_y)$, translation $t_c = (x_c, y_c, z_c)$, and rotation matrix $R_c$. 
The Jacobian of this camera is
\begin{equation}
    J = \begin{bmatrix}
    \frac{f_x}{z_c} & 0 & -\frac{f_x x_c}{z_c^2} \\[3mm]  % use extra space so rows don't overlap
    0 & \frac{f_y}{z_c} & -\frac{f_y y_c}{z_c^2}
    \end{bmatrix}.
\end{equation}
From the camera perspective, the 2D covariance of the Gaussian is $\Sigma_{proj} = J R_c \Sigma R_c^T J^T$.
Let $\mu_p$ represent the translation of a Gaussian in pixel space.

After sorting the Gaussians based on their depth and their visibility to each pixel, the Gaussians are rasterized onto an image as follows.
The opacity $\alpha$ of a Gaussian $k$ to a pixel located at $\rho = (\rho_x, \rho_y)$ units from the image center is
\begin{equation}
    \alpha_k = \delta ~ exp(-\frac{1}{2} (\mu_\rho - \rho)^T \Sigma_{proj}^{-1} (\mu_\rho - \rho)).
\end{equation}
The final color of pixel $\rho$ is computed as 
\begin{equation}
    C = \sum_{i=1}^{N} \alpha_i \prod_{j=1}^{i-1} (1 - \alpha_j) c_i,
\end{equation}
where $N$ is the number of Gaussians visible to pixel $\rho$, and the view-dependent color $c_i$ is computed using spherical harmonics.

\section{Method}
To produce a mobile-ready 3D Gaussian Avatar, we first train a high-fidelity avatar driven by Linear Blend Skinning (LBS) and  nonlinear correctives computed by a 2D convolutional decoder (Section~\ref{sec:compact_3d_gaussian_avatar}). 
This decoder is then distilled into a linear model that can run efficiently on mobile hardware with minimal quality degradation (Section~\ref{sec:linearization}).

\subsection{Compact 3D Gaussian Avatar}
\label{sec:compact_3d_gaussian_avatar}

\begin{figure*}[h]
    \centering
    \includegraphics[width=0.8\textwidth]{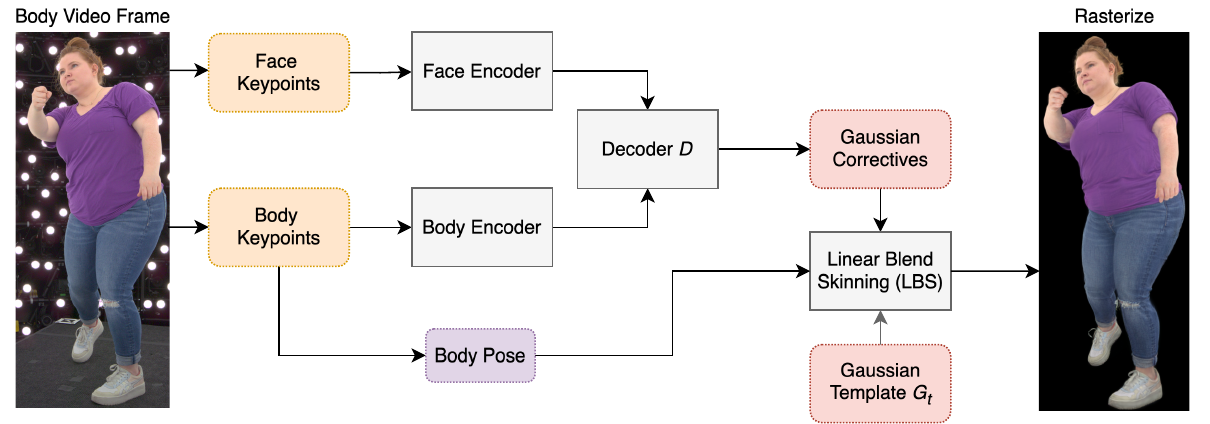}
    \caption{\textbf{System Diagram during training.} This is the configuration we use for training the model in Section~\ref{sec:compact_3d_gaussian_avatar}.}
    \Description{System Diagram}
    \label{fig:system_diagram_during_training}
\end{figure*}

Our high-fidelity 3D avatar consists of a convolutional decoder that predicts a 2D gaussian map $\mathrm{G}\in\mathbb{R}^{K\times256\times256}$ where each pixel represents a Gaussian and $K=37$ are the degrees of freedom of each Gaussian. 
In contrast to Animatable Gaussians~\cite{AnimatableGaussians}, which uses front/back orthogonal projection with many unused pixels, we project Gaussian properties to the UV map of a skinned body mesh. This results in a more compact representation where we can achieve higher Gaussian counts with smaller 2D maps. 

\paragraph{Initialization} 
To ease the training, we use a personalized skinned mesh as our starting point. 
We obtain this personalized mesh for each identity using LBS-based surface tracking ~\cite{Gall2009}, which also estimates the body poses required for training. 
%followed by Laplacian deformation to 3D scans ~\cite{Sorkine2004, Botsch2008}. 
To initialize the Gaussians, we project the positions and skinning weights of the mesh to the UV domain. After initialization, the mesh is no longer used.
 
\paragraph{Animation} We drive the animation of the avatar using LBS and non-linear correctives computed from a face code $\mathrm{e_f}\in\mathbb{R}^{32}$, a body code $\mathrm{e_b}\in\mathbb{R}^{32}$. At training time, these codes are estimated from face and body keypoints respectively, which are processed by two separate MLP encoders (see Figure \ref{fig:system_diagram_during_training}). 
The 2D Gaussian map is then computed as follows:

\begin{equation}
    G = G_t + M(D(e_f, e_b))
    \label{eqn:decoder}
\end{equation}

where $\mathrm{G_t}$ is a learnable template and $D$ is the convolutional decoder that estimates non-linear correctives for the Gaussian positions, rotations, scales and spherical harmonics, in canonical space. 
And, $\mathrm{M}$ is a binary mask that removes Gaussians that align with dead space in the UV domain, reducing the number of Gaussians from 65536 to 60381.
We do not apply correctives to the Gaussian $\delta$ (i.e. the term used to calculate opacity), since using constant $\delta$ greatly simplifies the pruning of non-visible Gaussians. 
The final step consists in converting the 2D Gaussian map $\mathrm{G}$ into 1D arrays of Gaussian properties $\{\mu_{c}, \Sigma_{c}, \delta, SH\}$ and applying LBS using the same formulation as in \cite{AnimatableGaussians},
\begin{equation}
\begin{split}
    \mu = \mathrm{R}\mu_{c} + \mathrm{t} \\
    \Sigma = \mathrm{R}\Sigma_{c} \mathrm{R}^\top 
    \label{eqn:lbs}
\end{split}
\end{equation}

where $\mathrm{R}$ and $\mathrm{t}$ are the blended joint rotations and translations, using the skinning weights assigned to each Gaussian at initialization.

\paragraph{Loss Functions.}
Our loss function is a weighted-sum of several terms:
$\mathcal{L}_{photo}$ and $\mathcal{L}_{lpips}$ are the photometric (L1) and LPIPS~\cite{LPIPS} distance between the ground-truth and predicted image.
$\mathcal{L}_{kpt}$ is the L1 loss, and the body pose keypoints.
To reduce "runaway" Gaussians that drift far from their initial position, $\mathcal{L}_{offset}$ is an L1 loss that is maximized when the mean $\mu$ of each Gaussian in $G$ is unchanged from $\mu$ where the Gaussian was initialized at the start of training.
To reduce unwanted holes in the avatar, we introduce $\mathcal{L}_{\alpha}$, which is an L1 loss that is maximized when the $\alpha$-transparency map generated by the model matches the Sapiens~\cite{sapiens_2024} segmentation mask; the intuition is that the avatar should be opaque and background pixels should be transparent.
When rendering Gaussians on a mobile, it is cheaper if each Gaussian is visible from fewer pixels~\cite{radsplat_2024}.
To reduce the number of pixels to which Gaussian is applied during rendering, we adopt the loss $\mathcal{L}_{\sigma}$, which uses L1 to minimize the average size of the Gaussians, and we adopt $\mathcal{L}_{opacity}$, which uses L1 to minimize the average opacity of the Gaussians. 
We visualize the effects of the losses in the supplementary material.
The weights for the sum are set to 
$\lambda_{photo} = 1$, 
$\lambda_{lpips} = 0.1$, 
$\lambda_{opacity} = 0.01$, 
$\lambda_{scale} = 1$, 
$\lambda_{offset} = 1$, 
$\lambda_{alpha} = 0.1$, 
and $\lambda_{kpt} = 0.1$.
The loss is computed as 
\begin{align}
\begin{split}
    \mathcal{L} = & \, \lambda_{photo} \mathcal{L}_{photo} + \lambda_{lpips} \mathcal{L}_{lpips} + \lambda_{\alpha} \mathcal{L}_{\alpha} + \lambda_{kpt} \mathcal{L}_{kpt} \\
    & + \lambda_{opacity} \mathcal{L}_{opacity} + \lambda_{scale} \mathcal{L}_{scale} + \lambda_{offset} \mathcal{L}_{offset} 
\end{split}
\end{align}

\paragraph{Implementation Details.}
The face and body MLP encoders consist of 2 hidden layers of size 256 followed by LeakyReLU activations. For better generalization, we apply the inverse head transform to the face keypoints and the inverse root transform to the body keypoints. 

The convolutional decoder $D$ begins with a linear layer that projects the input code to $32\times4\times4$, followed by 6 convolutional blocks that produce a $K\times256\times256$ Gaussian corrective map. 
Each convolutional block consists of a bilinear upsampling operation followed by two convolutional layers and LeakyReLU activations, and skip connections.
The decoder, encoders, and template $\mathrm{G}_t$ are trained end-to-end for 300k iterations, which takes around 20h on an NVIDIA A100 GPU.

\subsection{Mobile-ready Distillation}
\label{sec:linearization}
\paragraph{Linear Correctives}
We now turn our attention distilling the decoder $D$ into linear layers using principal component analysis (PCA).
For short, we refer to this process as {\em linearization}.
To distill a decoder for one identity, we begin by collecting a dataset of inputs and outputs to the encoder-decoder.
In this section, we continue to use the face-encoder, but we replace the body-encoder with LBS data.
So, our input space is a set of poses, and the output space is the output of the decoder from Equation~\ref{eqn:decoder}.

For one frame, the linearized decoder's input is a pose vector $p$, which the concatenation of the face encoder's embedding and the body pose in quaternion form.
We use PCA to compress the input vector $\theta$ to a total of 64 dimensions, which we then augment with a columns of ones to form a design matrix $C$.
We compute the linear layer in the least-squares sense using the normal equations i.e., we form $(C^TC)^{-1}C^T$ as the pseudo-inverse of the design matrix $C$ which is then multiplied by the decoder's output.
We describe the linear distillation more precisely in Algorithm~\ref{algo:distillation}.

\algrenewcommand\algorithmicrequire{\textbf{Input:}}
\algrenewcommand\algorithmicensure{\textbf{Output:}}
\begin{algorithm}
\caption{Distilled Decoder Correctives Computation}
\label{algo:distillation}
\begin{algorithmic}[1]
\Statex \hspace{-2em} \textbf{Stage 1: Calculate Bases}
\Require Set of input poses \(\{p_i\}\), Masked Correctives \(\{M(D(p_i))\}\)
\Ensure Pose Basis \( B_{p} \), Correctives Basis \( B_c \), Mean Pose \(\bar{p} \)
\State \(\bar{p} \gets \text{mean}(\{p_i\})\)
\State \( B_{p} \gets \text{PCA}(\{p_i - \bar{p}\}) \)
\State \( C \gets \{[1, (p_i - \bar{p}) \cdot B_{p}]\} \)
\State \( B_c \gets (C^T C)^{-1} C^T \cdot \{M(D(p_i))\} \)
\State \Return \( \bar{p}, B_{p}, B_c \)
\Statex \hspace{-2em} \textbf{Stage 2: Calculate Correctives for New Pose}
\Require Pose Basis \( B_{p} \), Correctives Basis \( B_c \), Mean Pose \(\bar{p} \), Novel Pose \( p \)
\Ensure Distilled Decoder Output \( D_L(p) \)
\setcounter{ALG@line}{0} % Reset line numbering
\State \( c \gets (p - \bar{p}) \cdot B_{p} \)
\State \( D_L(p) \gets [1, c] \cdot B_c \)
\State \Return \( D_L(p) \)
\end{algorithmic}
\end{algorithm}

The distilled decoder is a 2-layer model.
The first layer is a linear layer that takes a 64d compressed vector and outputs a 60381x16 vector, and the layer has 64x60381x16 parameters.
The second layer takes 6 of the 16 values from the first layer and expands them into 27 spherical harmonics, and the layer has 6x27 parameters. The remaining 10 channels are used for the Gaussian geometric parameters such as rotation, scale, and translation.
By far, the dominant cost is the first layer, which we find takes 5 ms to run (with quantization) on the neural processing unit (NPU) of the XR2G2 chip in the Meta Quest 3 VR headset.

With $D_L$ representing the distilled decoder, we compute the Gaussians for one frame of animation as follows.

\begin{equation}
    G = LBS(G_t + D_L(p), \theta)
    \label{eqn:linear}
\end{equation}

\paragraph{Corrective Sharing.}
\label{sec:gcs}

\begin{figure*}[h]
    % https://app.diagrams.net/#G1xJlLTWFSO8dkA2I4P4-iTofWz1eJoLVT#%7B%22pageId%22%3A%22RpzhsXeU8CXLvY0-wUcC%22%7D
    \centering
    \includegraphics[width=0.8\textwidth]{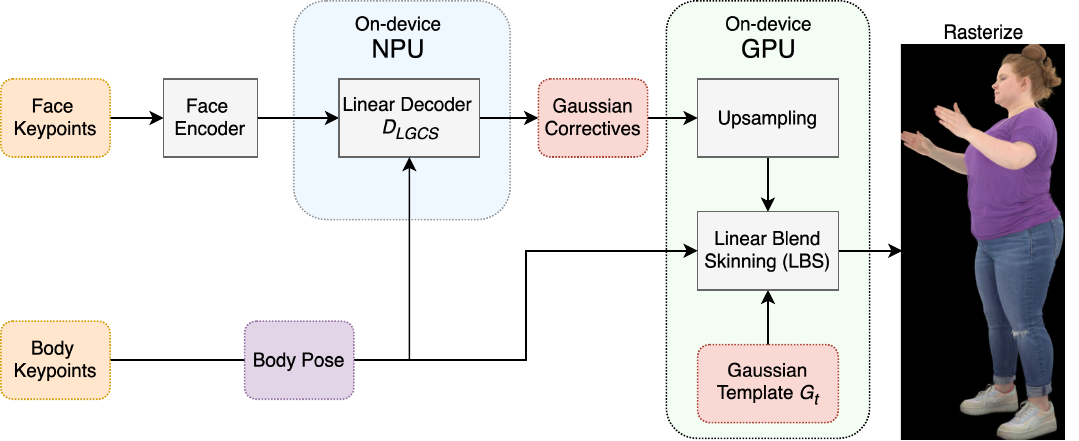}
    \caption{\textbf{System Diagram with optimizations.} We show the end-to-end optimized system, including the techniques from Sections~\ref{sec:linearization}.}
    \Description{System Diagram}
    \label{fig:system_diagram_with_optimizations}
\end{figure*}

While the static components of Gaussians need to be high-frequency, nearby particles of skin, hair, and clothing move together, making the displacements mostly low-frequency. Based on this observation, we propose to reduce the degrees of freedom in the correctives. More specifically, we group nearby Gaussians together and let them share the correctives. This effectively reduces the required rows of a linear matrix, leading to significant memory and computational time reduction on mobile.

While the decoder from Equation~\ref{eqn:decoder} produces a size 256x256x37 output, we modify the decoder to produce a smaller 64x64x37 output.
For notation, we call the modified decoder $D_{GCS}$, where GCS is short for Gaussian corrective sharing.
This reduces the number of correctives from 65536 to 4096. 
During training, we take the output of $D_{GCS}$ and use $\texttt{Nearest}$ interpolation to upscale it from 64x64 to 256x256 = 65536 correctives.\footnote{We experimented with both bilinear interpolation and nearest interpolation. See supplementary material for details.}
This has the effect of "sharing" correctives for each 4x4 neighborhood of Gaussians in UV-space.
For notation, we abbreviate $\texttt{Nearest}$ upscaling as $Up$.
During training, we compute the Gaussians for one frame of animation as
\begin{equation}
    G = LBS(G_t + M(Up(D_{GCS}(e_f, e_b))), \theta)
    \label{eqn:corrective_sharing}
\end{equation}
Finally, to combine linear distillation and corrective sharing, we compute the Gaussians as
\begin{equation}
    G = LBS(G_t + M(Up(D_{LGCS}(p))), \theta)
    \label{eqn:LGCS}
\end{equation}
We illustrate the end-to-end system with $D_{LGCS}$ and on-device inference in Figure~\ref{fig:system_diagram_with_optimizations}.

\subsection{Rendering}
\label{sec:visualizer}

We have implemented our visualization system in Vulkan to take advantage of the hardware rasterization for accelerated Gaussian splatting, as in~\cite{fast_gauss, Niedermayr_2024_CVPR}.
This is crucial for achieving high performance in compute-constrained setups like mobile devices. 
We now provide a brief overview of the rendering pipeline in VR. 
We first run a compute step that animates the avatars with linear blend skinning as proposed in Fig. 2. 
Next, we perform per-primitive culling to remove Gaussians that are outside the field-of-view for both stereo views in VR. 
Then, we project and sort the remaining Gaussians according to their view depth and reuse the sorted indices across both eye views. 
Finally, we expand the projected Gaussians into quad primitives with corresponding colors and opacities, which are rasterized and blended via a traditional graphics pipeline. 
The rationale for this final step is that Gaussian primitives are ellipsoids, but traditional graphics hardware is optimized for triangles and quadrilaterals.
All of the stages are implemented in standard Vulkan compute shaders except for the last (rasterization) stage, which uses a combination of a vertex and a fragment shader.

\begin{figure}[h]
    \centering
    \includegraphics[width=0.4\textwidth]{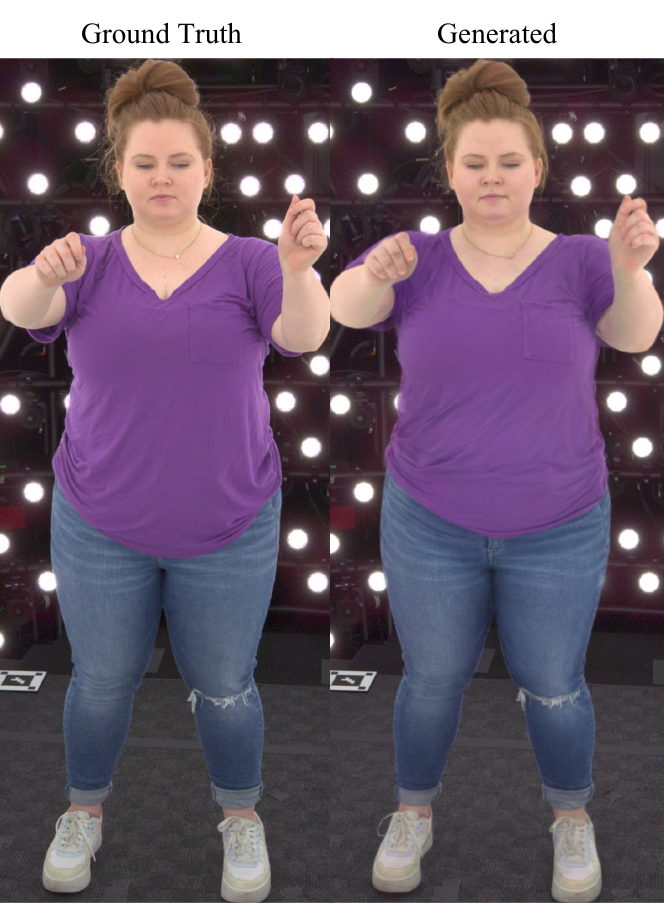}
    \caption{Qualitative comparison of the mobile-ready avatars. Left: ground-truth image. Right: generated image, with the avatar rendered in front of a static image of the studio. Both images are cropped to a bounding-box of a human segmentation mask.}
    \Description{Grid of avatars}
    \label{fig:how_we_evaluate}
\end{figure}

\section{Experimental Setup}

\begin{table*}[]
\caption{\textbf{Main results.} We evaluate the impact of linearization and number of correctives on quality. Results are at the original image resolution, with images cropped to rectangles based on segmentation of the avatar. Results are averaged over 4 identities. Latency is for the decoder only. One of the identities in Linear from scratch diverged due to instability and the reported number is the average across the remaining 3 identities.}\label{tab:5_id_results}

{
\begin{tabularx}{0.9\textwidth}{lccccccc}
    \hline
    Model & \multicolumn{1}{l}{\# Gaussians} & \multicolumn{1}{l}{\# Correctives} & \multicolumn{1}{l}{L1  $\downarrow$} & \multicolumn{1}{l}{LPIPS $\downarrow$} & \multicolumn{1}{l}{PSNR $\uparrow$} & \multicolumn{1}{l}{SSIM $\uparrow$} & \multicolumn{1}{l}{\begin{tabular}[c]{@{}c@{}}Decoder Latency \\ on Quest 3\end{tabular}} \\
    \hline
    {\small Animatable Gaussians~\cite{AnimatableGaussians}} & 300k & 300k  & 0.037 & 0.143 & 24.637 & 0.620 & \\
    SqueezeMe (initial) & 60k & 60k & 0.036 & 0.146 & 24.992 & 0.638 & \\
    SqueezeMe (GCS) & 60k & 4k & 0.036 & 0.147 & 25.024 & 0.636 & \\
    SqueezeMe (Linearized) & 60k & 60k & 0.039 & 0.150 & 24.411 & 0.632 & 5.0 ms \\
    SqueezeMe (GCS + Linearized) & 60k & 4k & 0.039 & 0.151 & 24.370 & 0.629 & 0.45 ms \\
    SqueezeMe (no decoder) & 60k & 0 & 0.040 & 0.164 & 23.905 & 0.622 & 0 \\
    Linear from scratch & 60k & 60k & 0.044* & 0.215* & 23.454* & 0.616* & 5.0 ms

\end{tabularx}
}
\end{table*}

We train and evaluate on data that was collected on an internal capture dome with the users' permission to use in published research. 
Our capture dome is a 3 meter diameter dome with 512, 25 megapixel cameras streaming at 90 FPS. 
The dome also has 1024 individually controllable lights.

For evaluating the results, we render the avatars in front of an image of the multi-camera, multi-light studio environment where the ground-truth data was captured, and we show examples of ground-truth and generated images in Figure~\ref{fig:how_we_evaluate}.
A significant portion of the image is background, so we crop the rendered images to the maximum width and height of Sapiens~\cite{sapiens_2024} segmentation masks on the ground-truth data.
It is important to note that without this cropping, all models would benefit from artificially inflated accuracy scores due to the large background regions that do not contain the avatar.
We use LPIPS~\cite{LPIPS}, L1, PSNR, and SSIM~\cite{SSIM} to evaluate the cropped images against ground-truth.

Training and evaluation use two separate sets of poses from distinctive animation segments.
All results tables, figures, and videos are computed on the evaluation set.
Further, in all of our evaluations, we are performing novel-view synthesis, i.e. we are using camera positions that were not in training set.
The evaluation set is held-out from the dataset that is used for training, linear distillation, and quantization.
For each subject, we use 2000-3000 frames per subject for training and 2000 held-out frames for evaluation.
For distillation, we use 512 frames from the training set.
All results are averaged across 4 identities and 5 cameras per identity.

For on-device demonstration, we use a Meta Quest 3 VR headset.
The Quest 3 is equipped with a Snapdragon XR2 Gen 2 chip, which integrates a CPU, a GPU, and a neural processing unit (NPU)~\cite{meta_quest_3}.
Unlike desktop GPUs, Snapdragon does not support CUDA; however, its mobile GPU is programmable via the Vulkan API, and its NPU can execute quantized neural networks exported from frameworks like PyTorch.

\begin{figure*}[h]
    \centering
    \includegraphics[width=0.95\textwidth]{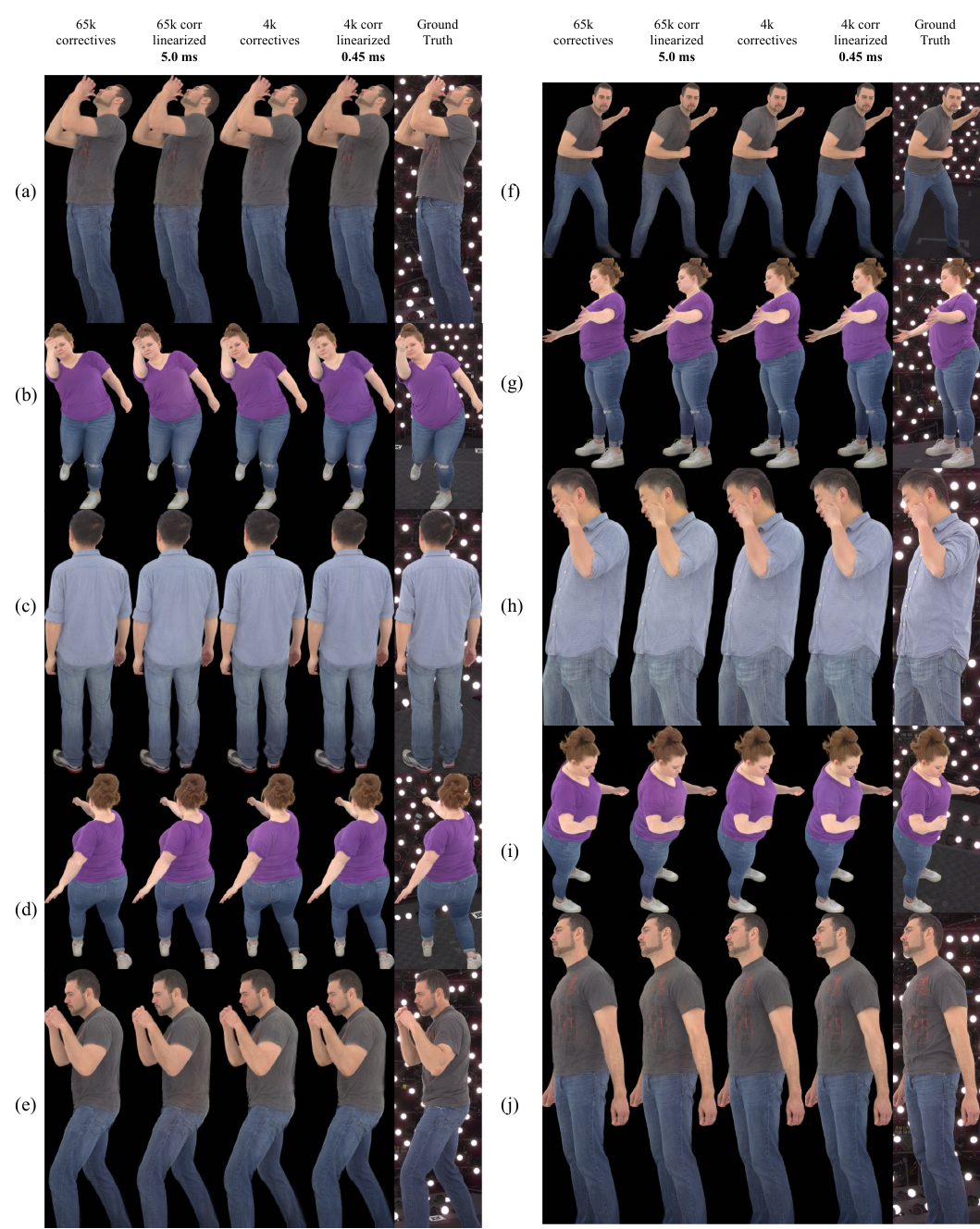}
    \caption{\textbf{Qualitative results.} Our 0.45 ms model produces results that are competitive with far more expensive models.}
    \Description{Grid of avatars}
    \label{fig:good_results}
\end{figure*}

\begin{figure*}[h]
    \centering
    \includegraphics[width=\textwidth]{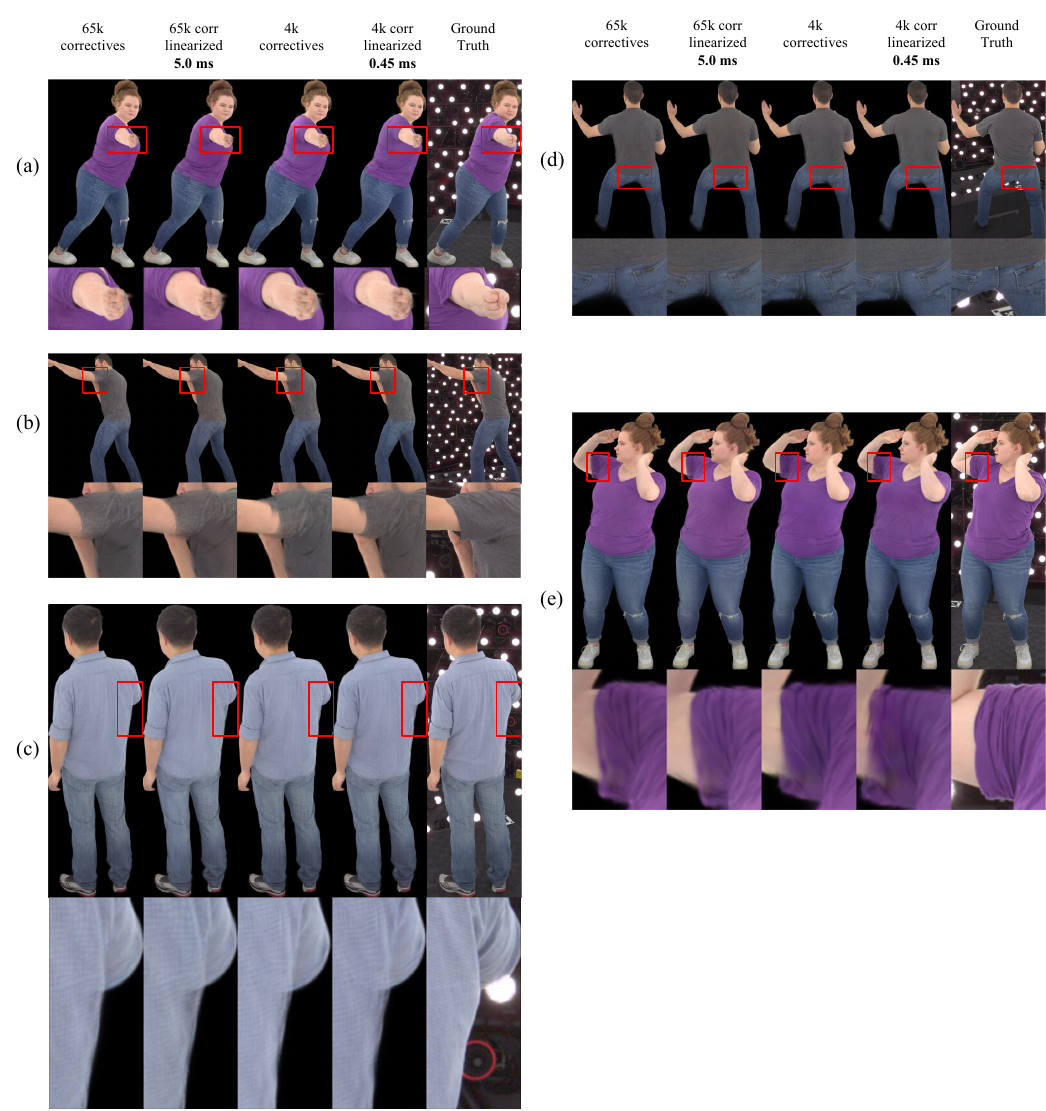}
    \caption{\textbf{Failure cases.} (a) In all models and identities, hands are sometimes blurry. (b) The 4k and linearized models struggle with the edge of a t-shirt sleeve. (c) All models have unwanted transparency under the arm for this identity's avatar, but it is worse in 4k and linearized models. (d) All models struggle with the seat of the pants for this identity, but the 4k and linearized models struggle more. (e) The 4k and linear models suffer more degradation in the underarm for this identity.}
    \Description{Grid of avatars}
    \label{fig:bad_results}
\end{figure*}

\section{Results}
\label{sec:results}

\paragraph{On-device Demo.}
In Figure~\ref{fig:teaser}, we illustrate 3 avatars running concurrently on a VR headset.
The decoder and Gaussian visualization run locally on the headset.
While we focus our evaluation on the Quest 3 headset, many smartphones and VR headsets have similar Qualcomm Snapdragon processors, indicating that similar results are achievable on a broad variety of mobile devices.
We provide a VR demo video in the supplementary material.

\paragraph{Quantitative Comparison.}
In Table~\ref{tab:5_id_results}, we compare several versions of our method with the baseline of Animatable Gaussians (AG)~\cite{AnimatableGaussians}. 
AG has 300k Gaussians and can run at 10 FPS on an NVIDIA GPU, but we found it infeasible to run on a mobile device. 
Without Gaussian corrective sharing (GCS) or linearization, our initial SqueezeMe model has a convolutional decoder with 60k Gaussians and 60k correctives.
Quantatively, our initial SqueezeMe model matches the L1 of Animatable Gaussians; it slightly underperforms the LPIPS of AG; and it slightly outperforms the PSNR and SSIM of AG.
Thus, our initial model produces comparable visual quality to AG, while having 5x fewer Gaussians.
At the other end of the spectrum with lower latency but also lower quality, we propose a version of SqueezeMe that has no decoder and relies entirely on linear blend skinning (LBS) to animate the Gaussian avatar; in Table~\ref{tab:5_id_results} this model significantly degrades on all metrics compared to AG and the initial SqueezeMe model.

Driving a tradeoff between complexity and quality, we propose a version of SqueezeMe with GCS, which reduces the number of correctives from 60k to 4k and has a smaller convolutional decoder; this model matches the L1 of the initial SqueezeMe model but degrades slightly on LPIPS, PSNR, and SSIM, as seen in Table~\ref{tab:5_id_results}.
Finally, we consider SqueezeMe models with linearization; these provide a midpoint between the quality metrics of the initial model and the no-decoder model, with decoder latency as low as 0.45 ms, making the linearized models compelling for practical on-device applications.

To show the effectiveness of the two-stage distillation strategy, let us consider the alternative of training a linear model from scratch without distillation.
In particular, we initialize a linear model with random weights and train it with the 7 losses that are used in the main paper.
Table~\ref{tab:5_id_results} shows that the "linear from scratch" model produces results that are significantly worse than the our SqueezeMe model.
Note that for one identity, training a linear model from scratch diverged, even with some exploration of learning rate and other hyperparameters. This suggests that training from scratch is less stable than training with distillation.
These factors illustrates the benefit of distillation for producing high-quality and low-latency Gaussian avatars.

In the supplementary material, we also evaluate SqueezeMe on the AvatarRex dataset~\cite{zheng2023avatarrex}.

\paragraph{Quantization.}
For on-device inference, we quantized the linear models with 8-bit weights and 16-bit activations.
In particular, we used post-training quantization.
We found that floating-point and quantized linear models produce identical quantative results.
Therefore, L1, LPIPS, PSNR, and SSIM of the "Linearized" results in Table~\ref{tab:5_id_results} are for both quantized and unquantized models.
The latencies reported in the table are for the quantized and linearized models.

\paragraph{Qualitative Comparison.}
In Figure~\ref{fig:good_results}, we show representative examples of the different models across different identities, poses, and camera views.
We observe high quality even with significant motion in the arms, torso, and legs.
We observe that using corrective-sharing to reduce the number of correctives from 65k to 4k and linearizing the model produce very little degradation in the avatar quality.
However, by reducing the number of correctives and linearizing the model, we are able to squeeze the decoder onto a VR headset with just 0.45 ms of latency per inference.

\paragraph{Limitations.}
We visualize failure-cases in Figure~\ref{fig:bad_results}.
SqueezeMe models intermittently produce blurry hands (see Figure~\ref{fig:bad_results}(a)), and unwanted transparency (see Figure~\ref{fig:bad_results}(c)).
Further, our optimizations -- namely, using Gaussian corrective sharing to reduce the number of correctives from 60k to 4k, and linear distillation -- introduce some additional artifacts that appear intermittently during inference. 
For instance, in Figure~\ref{fig:bad_results}(b, e) we find that corrective-sharing and linearization can both cause degradation at on the arms, particularly at the armpit and the point where a t-shirt sleeve meets the skin.
Finally, in Figure~\ref{fig:bad_results}(d) we observe that corrective-sharing reduces the visual quality on the seat of the pants, especially when the avatar is standing with legs apart.
To offer an intuitive explanation: corrective sharing assumes that neighboring Gaussians move together, but the arm and leg joints Gaussians may move more independently, causing artifacts to sometimes appear in those regions.
These problems may be resolved in the future with more adaptive methods of distributing Gaussians and correctives across a human avatar.

\section{Conclusion}
We have proposed a system comprised of multiple techniques to improve the efficiency of Gaussian Splatting in animatable human avatars, including a compact 3D Gaussian avatar representation, linear distillation, and corrective-sharing.
This improves the latency of the Gaussian corrective decoder from a baseline of 50 ms to just 0.45 ms.
Further, we show that running 3 avatars at 72 FPS on a Quest 3 VR headset is now possible.

By drastically reducing computational costs, this work not only advances photorealistic avatar rendering in VR but also expands the possibilities for practical applications. 
These include lifelike telepresence for remote collaboration, enhanced realism in multiplayer VR environments, and greater inclusivity by making high-quality avatars feasible on mobile-grade hardware. 
Looking ahead, the techniques introduced here could serve as the foundation for scaling avatar systems to support larger numbers of participants and for exploring their integration with augmented reality systems, bridging the gap between physical and virtual worlds.

{
    \small
    \bibliographystyle{ieeenat_fullname}
    \bibliography{main}
}

\clearpage

\begin{center}
  \textsf{\Huge\bfseries Appendix}
\end{center}

\appendix
\section{Upsampling Method}
Our initial SqueezeMe decoder has 65k outputs, which are reduced to 60k correctives by masking. 
These 60k pose-dependent correctives are applied to the 60k Gaussians to improve their appearance.
With Gaussian corrective sharing (GCS), the decoder has 4k outputs, which we upsample to 65k and then mask to 60k.
In this section, we compare two methods for upsampling: Nearest and Bilinear interpolation.
We present the comparison in Table~\ref{tab:upsampling_comparison}.
The numerical results show that Bilinear has a small advantage in quality over Nearest upsampling.

\begin{figure*}[h]
    \centering
    \includegraphics[width=0.6\textwidth]{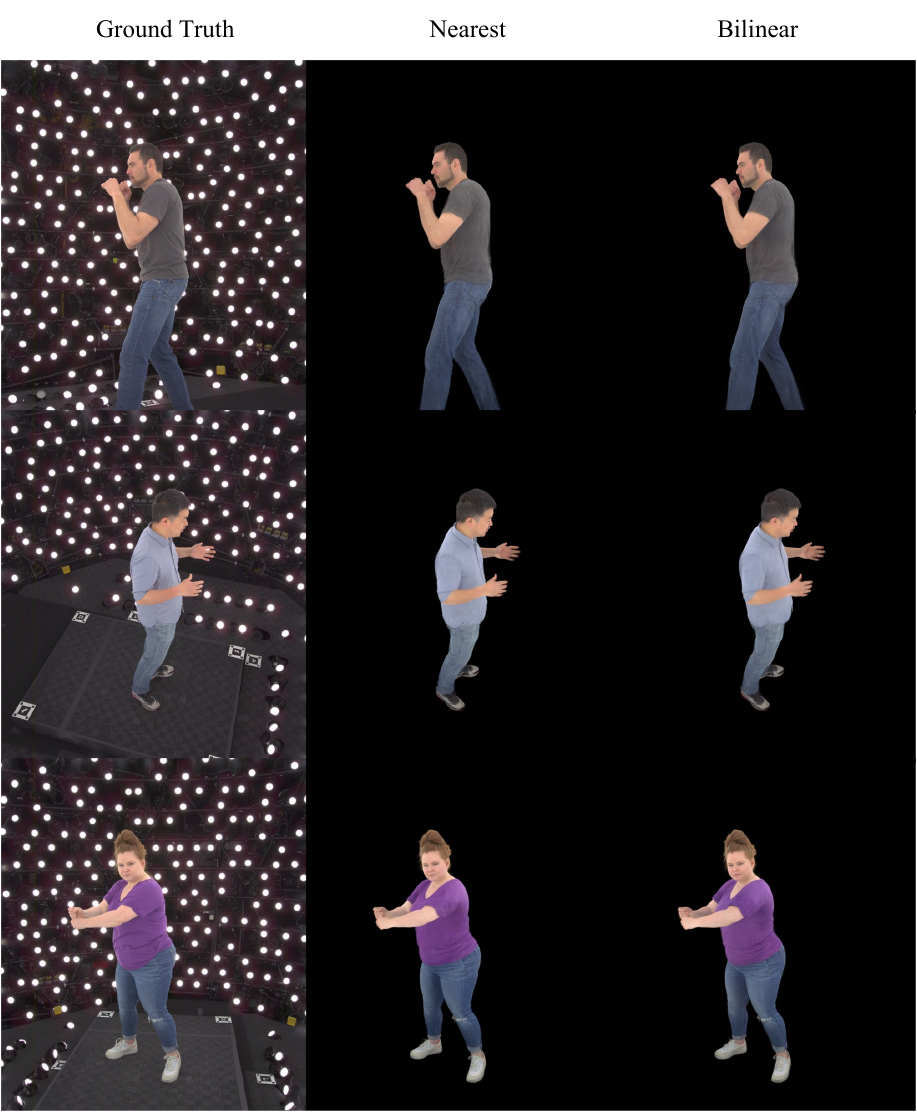}
    \caption{\textbf{Comparing Bilinear and Nearest interpolation in SqueezeMe (GCS).}}
    \Description{Grid of avatars}
    \label{fig:bilinear_vs_nearest}
\end{figure*}

Further, we provide a qualitative evaluation in Figure~\ref{fig:bilinear_vs_nearest} and find that the two methods produce very similar results.
For ease of implementation, and to reduce the load on the mobile GPU, we selected nearest interpolation for our VR demo videos and for the GCS results in the main paper.

\begin{table*}[b]
\caption{\textbf{Comparison of upsampling methods for Gaussian corrective sharing.} Results are averaged over 4 identities and are directly comparable to the Main Results table in the paper.}
\label{tab:upsampling_comparison}
\begin{tabularx}{0.85\textwidth}{lccccccc}
    \hline
    Model & Upsampling & \multicolumn{1}{l}{\# Gaussians} & \multicolumn{1}{l}{\# Correctives} & \multicolumn{1}{l}{L1  $\downarrow$} & \multicolumn{1}{l}{LPIPS $\downarrow$} & \multicolumn{1}{l}{PSNR $\uparrow$} & \multicolumn{1}{l}{SSIM $\uparrow$}  \\
    \hline
    SqueezeMe (GCS) & Nearest & 60k & 4k & 0.036 & 0.147 & 25.024 & 0.636   \\
    SqueezeMe (GCS + Linearized) & Nearest & 60k & 4k & 0.039 & 0.151 & 24.370 & 0.629    \\
    SqueezeMe (GCS) & Bilinear & 60k & 4k & 0.036 & 0.146 & 25.075 & 0.638 \\
    SqueezeMe (GCS + Linearized) & Bilinear & 60k & 4k & 0.039 & 0.150 & 24.399 & 0.630  \\
\end{tabularx}
\end{table*}

\section{Further Discussion of Loss Functions}

\begin{figure*}[h]
    \centering
    \includegraphics[width=0.85\textwidth]{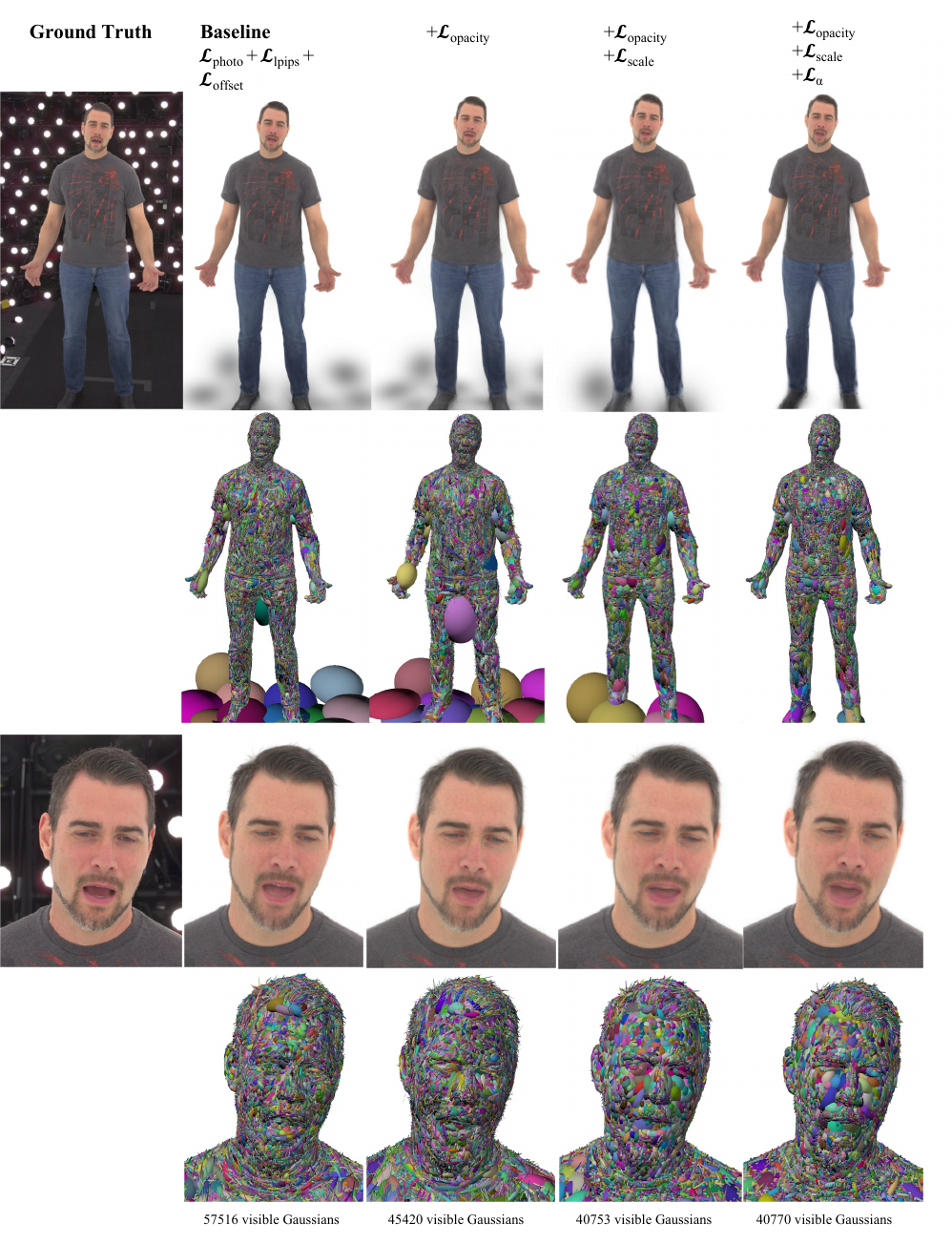}
    \caption{\textbf{Qualitative ablation study.} This figure evaluates the effect of the losses used to train the convolutional decoder. Using the losses of Animatable Gaussians \cite{AnimatableGaussians} as our baseline, we incorporate an opacity and scale loss to encourage the model to use less and smaller Gaussians. Prunning the non-visible Gaussians reduces the Gaussian count of the full-body avatar by 28\% with minimal quality loss. The alpha loss mitigates transparency issues and removes floaters around the avatar, greatly improving the final result.}
    \Description{Grid of avatars}
    \label{fig:ablation}
\end{figure*}

In the main paper, we trained with 7 loss functions: $\mathcal{L}_{photo}$, $\mathcal{L}_{lpips}$, $\mathcal{L}_{\alpha}$, $\mathcal{L}_{kpt}$, $\mathcal{L}_{opacity}$, $\mathcal{L}_{scale}$, and $\mathcal{L}_{offset}$.
In Figure~\ref{fig:ablation}, we illustrate the impacts of these losses.

\section{Results on AvatarRex Dataset}
So far, this paper and supplementary material have focused on an internal dataset that was captured with 512 cameras.
Now, we consider the AvatarRex dataset, which was proposed in~\cite{zheng2023avatarrex}.\footnote{Specifically, we use the AvatarRex data that is published at \url{https://github.com/lizhe00/AnimatableGaussians/blob/master/AVATARREX_DATASET.md}.}
The AvatarRex dataset has 4 identities and 16 cameras, with approximately 2000 frames per identity.
In particular, 14 camera views are used for training and 2 cameras are held-out for evaluation.
Further, for each avatar we hold out 500 frames per evaluation, therefore the training poses and evaluation poses are different.
Note that prior works such as \cite{AnimatableGaussians} use the same poses for training and evaluation.
For all our experiments on this dataset, we report results averaged across the identities zzr and lbn2.

\begin{figure*}[h]
    \centering
    \includegraphics[width=0.6\textwidth]{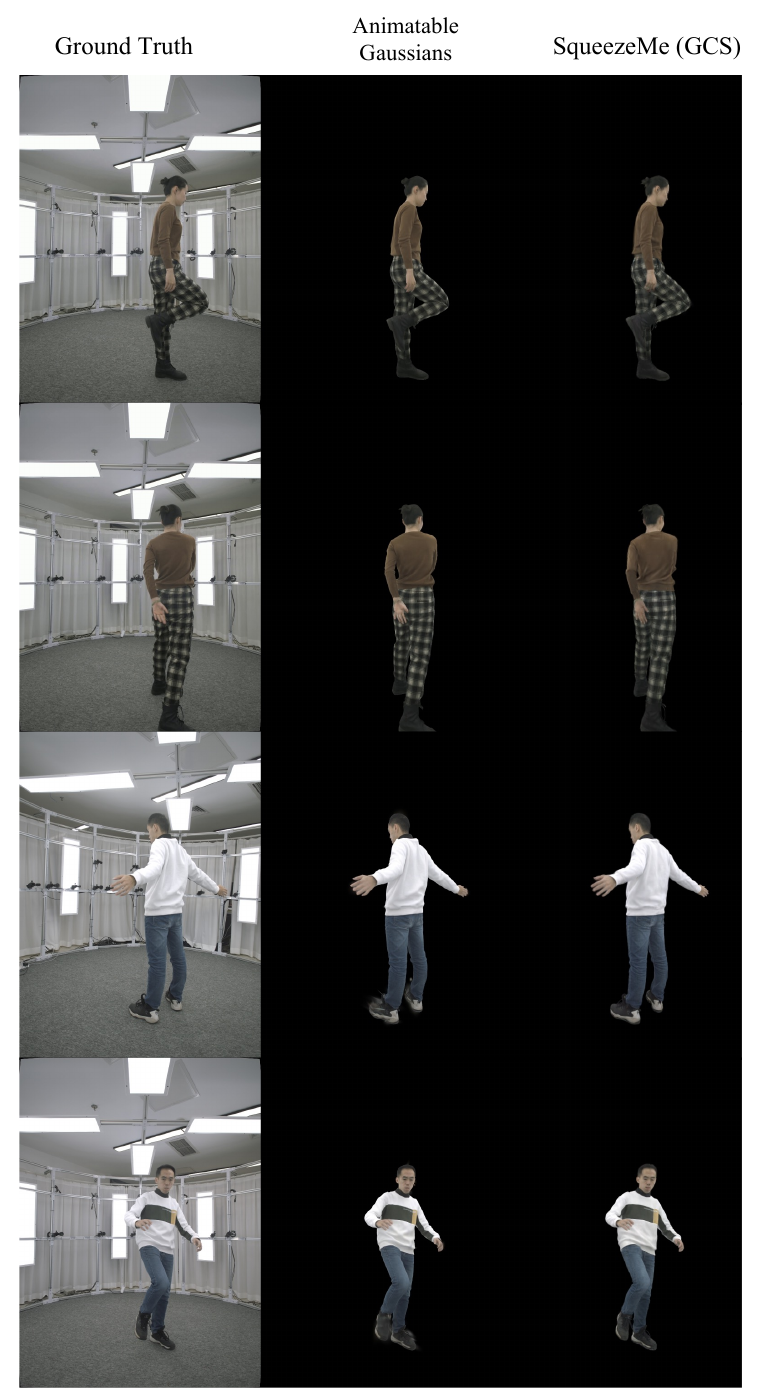}
    \caption{\textbf{Qualitative results on AvatarRex.}}
    \Description{Grid of avatars}
    \label{fig:avatarrex}
\end{figure*}

We present results on AvatarRex in Table~\ref{tab:avatarrex_results}.
When comparing Animatable Gaussians (AG) with 300k Gaussians to SqueezeMe with 60k Gaussians and 60k correctives, we observe that SqueezeMe outperforms AG on L1, PSNR, and SSIM.
Further, when comparing AG to comparing SqueezeMe with GCS and just 4k correctives, we find that SqueezeMe (GCS) outperforms AG on L1 and SSIM, and SqueezeMe (GCS) matches AG on L1.
Meanwhile, AG outperforms SqueezeMe on LPIPS.
Finally, we visually compare SqueezeMe and AG in Figure~\ref{fig:avatarrex}, and we observe similar quality for both methods.

\begin{table*}[b]
\caption{\textbf{Results on AvatarRex.} Results are averaged over two identities.}
\label{tab:avatarrex_results}
\begin{tabularx}{0.8\textwidth}{lcccccc}
    \hline
    Model & \multicolumn{1}{l}{\# Gaussians} & \multicolumn{1}{l}{\# Correctives} & \multicolumn{1}{l}{L1  $\downarrow$} & \multicolumn{1}{l}{LPIPS $\downarrow$} & \multicolumn{1}{l}{PSNR $\uparrow$} & \multicolumn{1}{l}{SSIM $\uparrow$}\\
    \hline
    Animatable Gaussians~\cite{AnimatableGaussians} & 300k & 300k & 0.059 & 0.151 & 19.542 & 0.844 \\
    SqueezeMe & 60k & 60k & 0.057 & 0.156 & 20.178 & 0.851 \\
    SqueezeMe (GCS) & 60k & 4k & 0.059 & 0.158 & 20.051 & 0.849 \\
\end{tabularx}
\end{table*}

\end{document}